\documentclass{article}

\usepackage[english]{babel}

\usepackage[letterpaper,top=2cm,bottom=2cm,left=3cm,right=3cm,marginparwidth=1.75cm]{geometry}

\usepackage{amsmath}
\usepackage{graphicx}
\usepackage[colorlinks=true, allcolors=blue]{hyperref}
\usepackage{array}
\usepackage{booktabs}
\usepackage{subcaption}

\graphicspath{{figures/}}

\newcommand{\tableref}[1]{Table~\ref{table:#1}}
\newcommand{\figref}[1]{Fig.~\ref{fig:#1}}

\newcommand{\dataset}{DPEB}

\begin{document}

\author{
  Md Saiful Islam Sajol$^{1}$, 
  Magesh Rajasekaran$^{1}$, 
  Hayden Gemeinhardt$^{1}$, \\
  Adam Bess$^{1}$, 
  Chris Alvin$^{2}$, 
  Supratik Mukhopadhyay$^{3,*}$ \\[1em]
  $^{1}$Computer Science, Louisiana State University, \\
  Baton Rouge, LA 70803, USA \\
  $^{2}$Computer Science, Furman University, Greenville, SC 29613, USA \\
  $^{3}$Environmental Sciences and Center for Computation and Technology, \\
  Louisiana State University, Baton Rouge, LA 70803, USA
}

\title{A Multimodal Human Protein Embeddings Database: DeepDrug Protein Embeddings Bank (\dataset)}

\maketitle

\abstract{
Computationally predicting protein-protein interactions (PPIs) is challenging due to the lack of integrated, multimodal protein representations. 
\dataset~is a curated collection  of 22,043 human proteins that integrates four embedding types: structural (AlphaFold2 \cite{AlphaFold2}), transformer-based sequence (BioEmbeddings \cite{bioembed}), contextual amino acid patterns (ESM-2: Evolutionary Scale Modeling \cite{esm}), and sequence-based n-gram statistics (ProtVec \cite{protvec}).
AlphaFold2 protein structures are available through public databases (e.g., AlphaFold2 Protein Structure Database), but the internal neural network embeddings are not.
\dataset~addresses this gap by providing AlphaFold2-derived embeddings for computational modeling.
Our benchmark evaluations show GraphSAGE with BioEmbedding achieved the highest PPI prediction performance (87.37\% AUROC, 79.16\% accuracy).
The framework also achieved 77.42\% accuracy for enzyme classification and 86.04\% accuracy for protein family classification.
\dataset~supports multiple graph neural network methods for PPI prediction, enabling applications in systems biology, drug target identification, pathway analysis, and disease mechanism studies.
\begin{center}
    \includegraphics[width=0.75\textwidth]{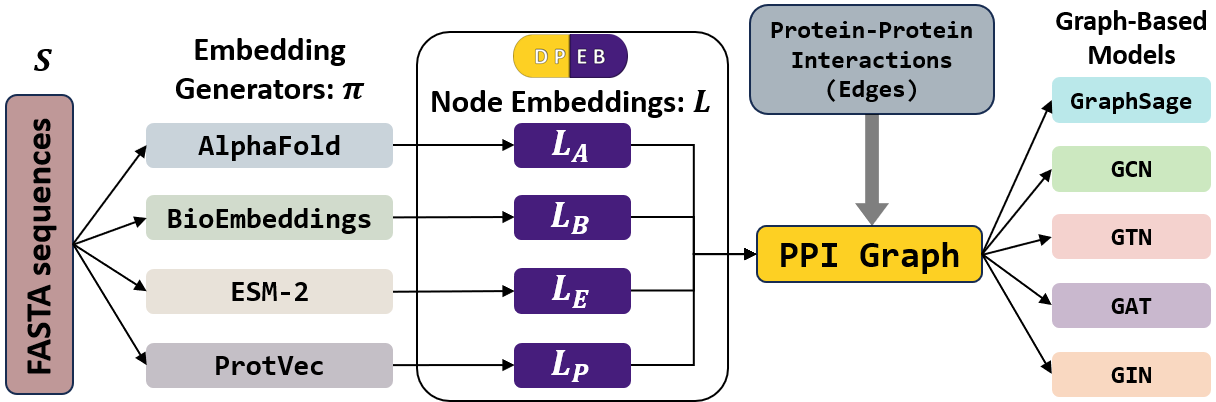}
\end{center}
}


\maketitle

\newcommand{\sm}[1]{{\bf $\spadesuit$ \textcolor{blue}{Dr. S: #1}}}
\newcommand{\ca}[1]{{\bf $\spadesuit$ \textcolor{cyan}{Alvin: #1}}}
\newcommand{\saiful}[1]{{\bf $\spadesuit$ \textcolor{violet}{Saiful: #1}}}

%
%

\section{Introduction}

Protein-protein interactions (PPIs) are essential for cellular functions, including complex protein formation and cellular organization.
The disruption of PPIs underlies numerous human diseases.

Experimental methods for studying PPIs are costly and slow when testing large numbers of protein pairs.
With protein data from multiple databases, \emph{in vivo} analysis of all potential PPIs is impractical.
Computational methods fill that gap by providing a cost-effective alternative that complements experimental approaches.

%
%
\section{Background}

\subsection{PPI Databases}
The Human Protein Reference Database (HPRD) \cite{keshava2009} contains approximately 41,327 human PPIs that have been manually curated from published literature.
In addition, it provides rich annotations detailing post-translational modifications, subcellular localization, and domain architecture.

BioGRID \cite{stark2006} offers a broader coverage having grown to over 1 million human PPIs since its initial release.
The Database of Interacting Proteins (DIP) \cite{salwinski2004} documents experimentally determined interactions, with 9,141 human interactions.
DIP includes a distinguished ``core'' subset for researchers requiring higher confidence data.

\begin{table*}[htbp]
\centering
\caption{Comparison of protein embedding and PPI databases.
\dataset~uniquely integrates four embedding types (AlphaFold2, BioEmbeddings, ESM-2, ProtVec) with $11$M human PPIs, while other databases provide single modalities or lack embedding-interaction integration.}
\renewcommand{\arraystretch}{1}
\resizebox{\textwidth}{!}{%
\begin{tabular}{>{\raggedright\arraybackslash}m{2.8cm}|>{\raggedright\arraybackslash}m{3.2cm}|>{\raggedright\arraybackslash}m{3.2cm}|>{\raggedright\arraybackslash}m{3.2cm}|>{\raggedright\arraybackslash}m{3.2cm}|>{\raggedright\arraybackslash}m{3.2cm}|>{\raggedright\arraybackslash}m{3.2cm}}
\toprule
\textbf{Dataset} & \textbf{Primary Data Type} & \textbf{Data Modality} & \textbf{Graph/Interaction Data} & \textbf{Intended Use Cases} & \textbf{Species} & \textbf{Coverage Info} \\
\midrule
DPEB & Multimodal Embeddings: AlphaFold2, BioEmbeddings, ESM-2, Protvec & Multimodal embeddings (latent space vectors) & Yes (11M PPIs from multiple sources) & Function prediction, PPI, graph-based architectures & Human-only ($\sim$22.4k proteins) & Human proteome (large-scale, up to 1,975 FASTA sequences) \\

\midrule
PCMol & AlphaFold2 latent embeddings for $\sim$4,331 proteins & Structural embeddings (latent space vectors) & No & De novo drug design, molecular generation & Multi-species ($\sim$4.3k) & 4,331 proteins (medium-sized, ligand data-rich subset) \\

\midrule
AlphaFold2 DB & Predicted 3D atomic structures (PDB/mmCIF files) & 3D structural coordinates and error estimates & No & Structural biology, visualization, modeling & Multispecies (incl. human) & $>$200M proteins from 47+ organisms \\

\midrule
STRING & Network-based protein embeddings from STRING PPI & Network topology embeddings (interaction-based) & Yes (predicted associations) & Cross-species function prediction, network analysis & Multispecies (12k+ organisms) & 1,322 eukaryotes, thousands of proteins/species \\

\midrule
BioPlex 3.0 & Experimental PPI interactions (AP-MS data) & Physical protein-protein interaction networks & Yes (experimental PPIs) & Interaction mapping, complex identification & Human (HEK293, HCT116 cells) &  118,162 interactions among 14,586 proteins, includes data from two human cell lines: \\
\bottomrule
\end{tabular}%
}
\label{table:refined_dataset_comparison}
\end{table*}

Context-specific interaction resources have also emerged.
The Integrated Interactions Database (IID) \cite{kotlyar2019} provides PPIs for 18 species across 133 tissues and 91 disease conditions, containing approximately 975,877 human PPIs (334,315 experimental and 667,804 predicted).
HIPPIE \cite{alanis2017} offers 273,900 experimentally detected interactions among 17,000 human proteins with filtering capabilities for tissue, function, and disease context.

BioPlex \cite{huttlin2015,huttlin2017} employs affinity purification-mass spectrometry to map the human interactome with 147,985 interactions among 12,695 proteins.
APID \cite{alonso2016} integrates PPIs from multiple primary databases with confidence scoring for over 400,000 annotated interactions.
Despite their diversity and comprehensiveness, these specialized databases lack precomputed protein embeddings for computational modeling.
\tableref{refined_dataset_comparison} compares databases that provide protein embeddings alongside interaction data.

%
%
\subsection{Computational PPI Prediction Methods.}
Early approaches applied traditional machine learning techniques to custom protein representations.
For instance, \cite{code4_knn} implemented k-nearest neighbors with optimized distance measures on protein sequence descriptors, achieving 85.15\% accuracy on yeast proteins but lacking evaluation on human proteins. 
Several works \cite{pcpip,ld_svm,mcd_svm,acc_svm} employed Support Vector Machines (SVMs) reaching accuracies of 88-91\% on yeast proteins.
Ensemble methods also showed promise, with \cite{lra_rf} utilizing rotation forest classifiers and \cite{gcforest} implementing deep forest approaches for PPI classification.
However, these approaches required manual feature engineering.

Deep learning methods automated feature extraction for PPI prediction.
DNN-PPI \cite{hang2018dnn} leverages deep neural networks to automatically learn protein representations from sequence-based descriptors, achieving an accuracy exceeding 94\%.
TAGPPI \cite{song2022tagppi} combines TextCNN with Graph Attention Networks to extract and fuse both sequence and structural features.
PIPR \cite{chen2019pipr} uses a Siamese residual RCNN approach to extract local features and contextual information.

Graph neural networks reformulate PPI prediction as a link prediction task.
GNN-PPI \cite{lv2021gnnppi} captures topological patterns for previously unseen interactions.
DL-PPI \cite{wu2023dlppi} combines Inception modules with attention mechanisms and a Feature Relational Reasoning Network.
GNNGL-PPI \cite{gnngl_ppi} uses global and local graph features with Graph Isomorphism Networks and protein embeddings derived from the MASSA model \cite{massa}.

Recently, PCMol~\cite{bernata2024pcmol} demonstrated the utility of AlphaFold2-derived structural embeddings by conditioning molecular generative models on them for de novo drug design.
They released a dataset of approximately 4{,}300 protein embeddings of different species, limited to sequences up to 1{,}536 amino acids.
However, existing approaches rely on single-modality protein representations, lack human proteome specificity, or cannot integrate complementary embeddings.

%
%
\subsection{Our Contribution}

We present \dataset, which integrates four embedding types for 22,043 human proteins: AlphaFold2 \cite{AlphaFold2} structural features, BioEmbeddings \cite{bioembed} sequence representations, ESM-2 \cite{esm} contextual patterns, and ProtVec \cite{protvec} n-gram statistics.
These complementary embeddings capture different aspects of protein structure and function.
The multimodal approach provides more comprehensive protein characterization than single embedding methods.

\dataset~enables comparison of predictions across multiple embedding types.
Thus, researchers can identify consensus predictions or method-specific interactions.
Different embedding types showed varying predictive accuracy, indicating each captures different protein properties.
BioEmbedding achieved the highest AUROC (87.37\%) in our experiments while AlphaFold2 captured structurally-driven interactions.
No single embedding type captured all protein properties, demonstrating the value of multimodal approaches.

\tableref{refined_dataset_comparison} shows that \dataset~combines human-specific data from UniProt \cite{uniprot}, STRING \cite{string}, and IntAct \cite{intact} with multimodal embeddings.
\dataset~supports multiple graph neural network architectures including GraphSAGE, GCNs, GTNs, GATs, and GINs.
This enables researchers to select appropriate methods for specific biological questions.

\sloppy

\section{Methods}

\begin{figure*}[t!]
    \centering
    \includegraphics[width=0.65\textwidth]{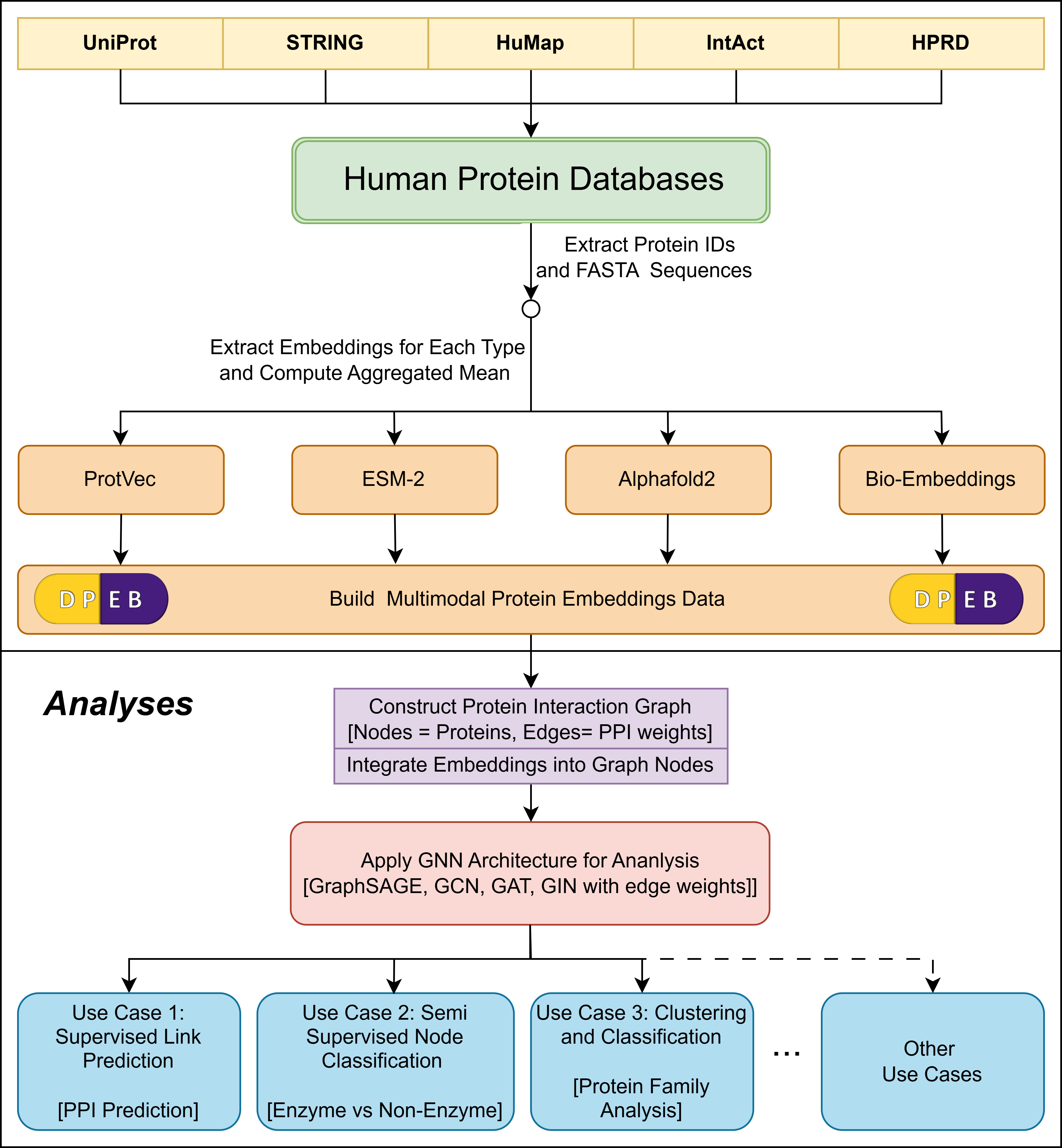}

    \caption{Overview of the DPEB pipeline for multimodal protein embedding analysis.
    Protein IDs and FASTA sequences are extracted from major human protein databases (UniProt, STRING, HuMap, IntAct, HPRD).
    Embeddings from four diverse sources---ProtVec, ESM-2, AlphaFold2, and BioEmbeddings---are computed and aggregated to build the \dataset (Deep Protein Embedding Benchmark) multimodal dataset.
    These embeddings can then be integrated into nodes of a protein-protein interaction (PPI) graph, where the edges denote interaction strengths represented by weighted connections. 
    Graph neural networks (GNNs), including GraphSAGE, GCN, GTN, GAT, and GIN with edge weight support, are applied to perform downstream tasks such as supervised link prediction (PPI), semi-supervised node classification (e.g., enzyme vs. non-enzyme), and protein family clustering and classification.}
    \label{fig:dpeb}
\end{figure*}

As shown in \figref{dpeb}, \dataset~can construct protein-protein interaction graphs where users can select different protein embeddings and incorporate interaction data from external sources such as HuMap.
Each protein functions as a node, with neighborhoods defined by interaction relationships.
Nodes carry features derived from amino acid sequences and structural properties, while edges encode residue interaction information from PPI databases.

%
%
\subsection{Data Sources and Integration}

Our framework integrates protein databases for sequence and interaction data.

%
%
\subsubsection{Primary Protein Sequence Data}
The \textbf{Universal Protein Resource (UniProt)}~\cite{uniprot} serves as our primary source for protein sequence data and functional annotations.
UniProt provides curated annotations on protein structure, biological functions, and post-translational modifications.
We leverage UniProt's manually curated Swiss-Prot entries to ensure reliability in protein feature extraction.
The database’s standardized nomenclature and cross-references enable integration with other datasets.

%
%
\subsubsection{Protein-Protein Interaction Data Sources}

The \textbf{Human Protein Reference Database (HPRD)} \cite{keshava2009} includes experimentally validated interactions between protein pairs, curated from peer-reviewed studies.
In addition to interaction data, it provides detailed biological annotations, including protein modifications, domain structures, and subcellular localization.
HPRD contributes a high-quality subset of human interactions that complement large-scale high-throughput datasets and prediction-based repositories.
Incorporating HPRD introduces manually reviewed interactions from published literature.

\textbf{STRING} (Search Tool for the Retrieval of Interacting Genes/Proteins)~\cite{string} is a widely used database that integrates both experimentally verified and computationally inferred protein-protein interactions.
It aggregates data from diverse sources, including large-scale experimental data, predictive algorithms, text mining of scientific literature, and manually curated pathway databases.

A distinctive feature of STRING is its assignment of confidence scores to protein associations, representing the estimated probability that the supporting evidence corroborates  the  biological relevance of an interaction.
It encompasses both direct molecular contacts and functional associations, providing broader context than databases focused solely on experimental binding evidence.
With coverage of over 14,000 organisms, STRING offers extensive taxonomic breadth while maintaining comprehensive documentation of human-human protein interactions.
This database provides annotations of both direct experimental evidence and interactions inferred from evolutionary conservation.

\textbf{IntAct} \cite{intact} is a freely accessible, open-source database that offers both molecular interaction data and integrated analysis tools.
Unlike STRING's scoring approach, IntAct focuses on manually curated, experimentally verified molecular interactions from literature or direct data submissions.
The curation model employed by IntAct prioritizes comprehensive annotation of interaction experiments, including the specific methodologies used, experimental conditions, and quantitative interaction parameters.
IntAct employs rigorous curation standards following International Molecular Exchange (IMEx) consortium guidelines, ensuring high data quality with comprehensive evidence trails.
The database supports complex interaction networks beyond simple binary interactions, representing molecular complexes and their hierarchical organization.

\textbf{HuMap} \cite{humap} represents a specialized, high-confidence human protein interactome map that addresses the quality limitations inherent in many PPI datasets.
HuMap integrates interactions detected by multiple orthogonal assays, substantially reducing false positives while maintaining extensive proteome coverage.
In contrast to general PPI databases, HuMap is dedicated solely to human binary interactions, which are validated using complementary experimental techniques.
The dataset employs standardized reliability scoring derived from experimental reproducibility metrics, allowing quality-based filtering for applications requiring different confidence thresholds.
We use HuMap as a source of accurate ground truth data for protein interactions.

%
%
\subsection{Protein Embedding Methodologies}

\dataset~provides four distinct protein embedding strategies: structural, sequence-based, contextual, and statistical. 
Users can select different embedding types based on their particular analysis requirements.

\subsubsection{Structural Embeddings}
\textbf{AlphaFold2} \cite{AlphaFold2} uses deep learning methods to predict protein structures.
It integrates multiple sequence alignments (MSAs), evolutionary information, and attention-based neural architectures to address the protein folding problem.
The system achieved accuracy comparable to experimental methods for many proteins at the 14th Critical Assessment of Protein Structure Prediction (CASP14).
AlphaFold2 generates structural embeddings that capture spatial relationships among amino acids in protein conformations.
These embeddings encode information about folding patterns, domain interactions, and potential binding sites.
AlphaFold2 translates sequence information into structural predictions that maintain physical and biochemical plausibility.

To obtain structurally informed embeddings, we employed a customized implementation of the AlphaFold2 pipeline configured to operate in single-sequence mode.
Specifically, we used the first model from the AlphaFold2 ensemble— Model 1 PTM, an optimized CASP14 monomer model that includes the predicted TM-score,  detailed in Sections 1.11.1 and 1.12.1 of the   Supplementary Information included with the original AlphaFold2 paper \cite{AlphaFold2}.
This model includes an additional head to predict the Template Modeling score (pTM), providing an estimate of global structural accuracy.
As one of the five official AlphaFold2 configurations used in the CASP14 competition, it serves as a robust monomeric structure predictor.
The model leverages template information and supports up to 5120 additional MSA sequences.

For each input protein sequence, AlphaFold2 was run in \textit{single-sequence mode} using only the first pre-trained model (\texttt{model\_1\_ptm}) from its ensemble of five.
This configuration bypasses the construction of multiple sequence alignments (MSAs), allowing the model to efficiently extract internal structural representations directly from individual sequences.
From the model output, we extracted intermediate residue-level embeddings known as the \textit{single representation}, a learned tensor that captures the contextual and structural characteristics of each amino acid position.
These per-residue vectors were aggregated and stored as fixed-length protein embeddings.
The resulting embeddings contain both structural and contextual (semantic) features, making them highly suitable for a variety of downstream computational tasks.

\subsubsection{Sequence-Based Embeddings}

\textbf{ProtVec}~\cite{protvec} uses natural language processing (NLP) techniques to transform protein sequences into continuous vector representations.
ProtVec conceptualizes proteins as sentences and amino acid n-grams as words, applying the skip-gram neural network model to learn relationships between local subsequences.
Unlike traditional methods that use predefined biochemical properties or structural information, ProtVec learns (unsupervised) representations directly from protein databases.

ProtVec created vector spaces where similar protein motifs cluster together and vector arithmetic reveals biochemical relationships.
Inspired by word2vec \cite{word2vec}, this approach transforms variable-length protein sequences into fixed-dimensional vectors (typically 100 dimensions).
These vectors are fed to machine learning models as  input features  for tasks such as protein classification, function prediction, and structural analysis.
ProtVec applies distributional semantics to protein sequences, capturing protein biochemistry without explicit feature engineering.

In our implementation, we utilized a skip-gram model pre-trained on the Swiss-Prot database, allowing us to capture local sequence patterns commonly observed in biologically relevant proteins.
For each protein, we computed embeddings using three offset tokenizations to account for different reading frames.
The embeddings from each offset were averaged and then concatenated to produce a 300-dimensional vector per protein.
This encoding strategy provides distributional representations by capturing local semantic patterns within overlapping k-mers.
We computed ProtVec embeddings for all proteins in the dataset using this approach.

%
%
\subsubsection{Language Model Embeddings}
 
\textbf{ESM-2} \cite{esm} embeddings are derived from a transformer-based model that is trained on billions of protein sequences using a self-supervised learning strategy.
They capture evolutionary context without MSA, and produce high-dimensional, residue-aware vectors suitable across diverse protein families.
To derive protein-level representations from evolutionary-scale data, we employed the pre-trained transformer model \texttt{esm2\_t33\_650M\_UR50D} from the ESM-2 family developed by Meta AI.
Each protein sequence was formatted into standard FASTA and processed through the ESM-2 embedding pipeline.
For every sequence, the model outputs token-level embeddings across 33 transformer layers.
We used the final layer (layer 33) and applied mean pooling across all residue embeddings to obtain a fixed-length, 1280-dimensional vector for each protein.
These embeddings capture both local and global contextual information, learned from large-scale corpora of protein sequences. 

\textbf{BioEmbeddings} \cite{bioembed} provides a standardized pipeline generating protein embeddings from sequences.
The framework implements various encoding strategies within a single infrastructure.
We generate protein embeddings using a \texttt{prottrans\_bert\_bfd} LM model \cite{elnaggar2021prottranscrackinglanguagelifes} trained on BFD large dataset \cite{steinegger2018clustering}.
This ProtTransBFD model uses BertTokenizer for tokenizing protein sequences. 

BioEmbeddings generates embeddings at different granularities: per-protein embeddings for global characterization, per-residue embeddings for fine-grained analysis, and specialized embeddings for specific prediction tasks.
The framework implements standardization procedures, ensuring vector representations maintain consistent dimensionality and numerical properties across different protein sequences without additional preprocessing.
These embeddings encode both the order and context of amino acids in protein sequences.

%
%
\subsubsection{Embedding Dimensionality and Normalization}
In the DPEB repository, we share both the raw (per-residue) and aggregated (per-protein) versions of each embedding.
The raw format retains the full contextual information captured by the models.
For example, ESM-2 provides a 1280-dimensional vector for every amino acid.
The aggregated version summarizes these using mean pooling to produce a fixed-length vector per protein (e.g., 1280-D for ESM-2, 384-D for AlphaFold2).

To preserve the integrity of the original models, we do not apply normalization or dimensional alignment across embedding types within the dataset.
As a result, the dimensions vary: ProtVec (300-D), BioEmbedding (1024-D), ESM-2 (1280-D), AlphaFold2 (384-D).
We chose not to enforce uniformity of dimensionality because different tasks and models may benefit from different preprocessing strategies.
Users can customize the data processing to their specific needs.
This includes normalizing the embeddings, projecting them into a common space, or using each representation in its original form.

%
%
\subsection{Graph Construction Methodology}
\label{sec:creation}

\dataset~is constructed by unifying four major human PPI repositories: HPRD, HuMAP, IntAct, and STRING.
Each dataset was preprocessed to retain only essential columns —namely the interacting protein pairs (source and target) and their associated interaction confidence scores (weight).

%
%
\subsubsection{Data Preprocessing and Standardization}
To ensure a consistent schema across datasets, column names were standardized and all entries were integrated into a single interaction list through concatenation.
The resulting dataset comprises over 11 million interaction pairs, providing extensive coverage of the human interactome while reflecting varying confidence levels across different data sources.

%
%
\subsubsection{Graph Construction}
A unified interaction graph was then constructed using the Deep Graph Library (DGL\footnote{https://www.dgl.ai/}), where each unique protein was mapped to a numeric node index.
Nodes in the graph were further enriched with high-dimensional protein embeddings derived from ProtVec, ESM, BioEmbedding, and AlphaFold2.
These embeddings encode semantic and structural information for each protein node.

Edge weights were mapped directly from the confidence scores in the unified dataset, normalized to the [0, 1] range, and incorporated into the graph as a \texttt{float32} tensor.
These weights allow graph neural networks to consider the biological confidence associated of each PPI.
Edges were filtered to include only those between proteins with valid embeddings.
Duplicate interactions were removed to avoid data leakage during train-test splits.
We saved the final graph and the protein-to-index mapping in Python pickle format for downstream analysis.

The detailed extraction processes and embedding pipelines are available in our public repository: \href{https://github.com/deepdrugai/DPEB}{DPEB}.

%
%
\subsubsection{Protein Sequence Characteristics}

\begin{figure}[t!]
    \centering
    \includegraphics[width=0.7\columnwidth]{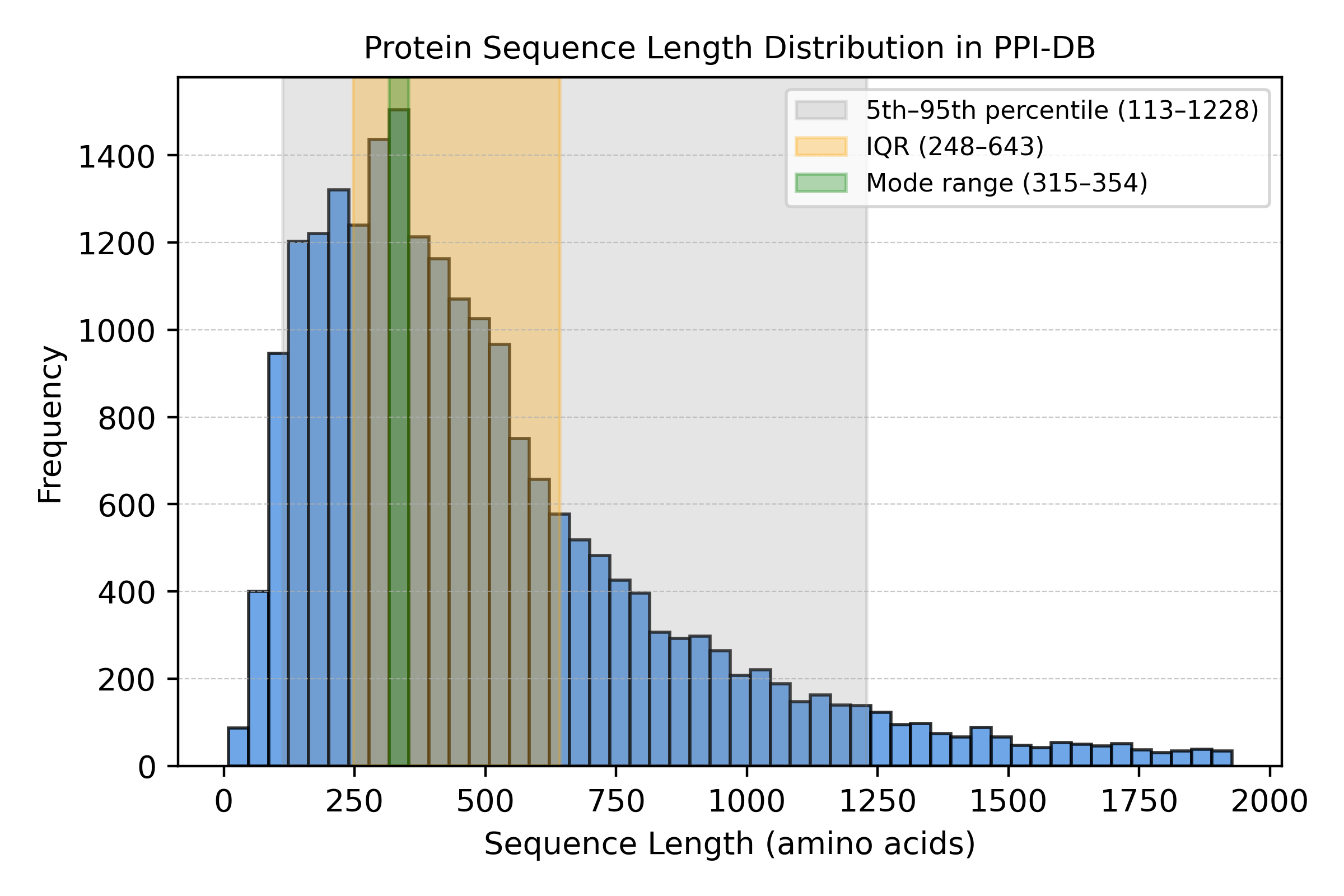}
    \caption{Distribution of human protein sequence lengths in the \dataset~. Most sequences range from 113 to 1228 residues (5th–95th percentile), with the highest frequency observed between 315 and 354 residues.}
    \label{fig:fasta_length}
\end{figure}

As depicted in \figref{fasta_length}, the distribution of human protein sequence lengths in our dataset reveals characteristic patterns.
The majority of sequences (90\%) fall within the 113–1228 residue range (5th–95th percentile), with the most frequent lengths concentrated between 316 and 354 amino acids.

%
%
\subsection{Benchmark Evaluation Framework}

\dataset~supports multiple graph-based neural network architectures for benchmarking protein interaction methods.

%
%
\subsubsection{Graph Neural Network Architectures}
We implemented and benchmarked five graph neural network (GNN) architectures: GraphSAGE, GAT, GCN, GTN, and GIN.
These architectures are commonly used for protein-protein interaction prediction and biological network analysis \cite{Ma2024Bingo,Xu2024SurveyPPIs, Kamp2025GNNsuite, Zhou2022ComparativePPI}.

We implemented the following architectures for protein interaction data:
\begin{itemize}
    \item \textbf{GraphSAGE}: A scalable graph neural network that efficiently creates node embeddings through local neighborhood sampling and feature pooling. 
    
    \item \textbf{Graph Attention Network (GAT)}: Uses attention mechanisms to weight neighbor contributions during node aggregation.
    
    \item \textbf{Graph Convolutional Network (GCN)}: Performs spectral convolution operations over nodes and their neighborhoods.

    \item \textbf{Graph Transformer Network (GTN)}: Combines self-attention mechanisms with graph structure to capture complex dependencies.
    
    \item \textbf{Graph Isomorphism Network (GIN)}: A powerful architecture designed to approximate graph isomorphism tests, enhancing the discriminative power of graph neural networks.

\end{itemize}

These architectures capture distinct aspects of protein interaction networks, from localized patterns to broader global graph structures.

%
%
\subsubsection{Model Evaluation Approach}
These baselines enable comparative analysis of different methods on human protein interaction prediction.
We provide three example applications using different protein embeddings from \dataset:
PPI prediction, functional classification, and structural analysis.

%
%

%
%
\section{Interactive Framework and Benchmark Applications}

\dataset~allows users to query protein embeddings using UniProt identifiers.
Users can select from the four embedding types:AlphaFold2, BioEmbeddings, ESM-2, and ProtVec.
The following subsections detail our three use cases.



\subsection{Use Case 1: Supervised Protein-Protein Interaction Link Prediction}

We predict PPIs using supervised link prediction problem on a human protein interaction graph with over 11 million known interaction pairs.
We evaluate the performance of multiple graph neural network (GNN) architectures with different protein embeddings.

%
%
\subsubsection{Experimental Setup}
We treat protein interaction prediction as binary edge classification.
The nodes depict proteins, while the edges represent known interactions with associated weights.
For each positive interaction, we generate one negative sample by randomly selecting a protein pair without observed interactions \cite{jha2022ppi,guan2025graphrpi,Zeng2024,Wu2023,Neumann2022,Chen2019}.
We exclude self-connections and maintain a 1:1 positive-to-negative ratio for balanced training.


We trained five GNN architectures—GraphSAGE, GAT, GCN, GTN, and GIN—to assess their effectiveness across different protein embedding modalities.
All models use DGL’s \texttt{NeighborSampler} for mini-batch training and binary cross-entropy loss for edge prediction.
Node features incorporate all four embeddings types from \dataset.

 The dataset was partitioned into training (70\%), validation (20\%), and testing (10\%) subsets using edge-level splitting with bidirectional interactions preserved.
We fixed the random seed to $42$ for reproducibility. 
The graph's large scale (millions of edges) made cross-validation infeasible, so we used a fixed train/validation/test split for all experiments.
To address the lack of cross-validation, we provide bootstrap evaluations in the supplementary material.
We trained all models for 20 epochs with batch size 512 utilizing the Adam optimizer on an NVIDIA Tesla V100-SXM2 (32GB) GPU using PyTorch and DGL.
Training took 7-8 hours per model.

We tested learning rates of \(\mathit{lr} = 10^{-3},\ 10^{-4},\ \text{and}\ 10^{-5} \) for all models and report the best results for each embedding type.
All models employed three graph convolution layers (or transformer layers for GTN) followed by an MLP-based edge predictor.
The predictor computed interaction probabilities using element-wise multiplication of source and destination node embeddings.

\begin{table}[t!]
\centering
\caption{Baseline F1, AUROC, and accuracy scores for GraphSAGE, GCN, GTN, GAT, and GIN using different protein embedding approaches on human protein datasets.
Models incorporate embeddings derived from AlphaFold2, ProtVec, ESM-2, and Bio Embeddings.
Models were not optimized.
Extended results are available in the Supplementary Section.}

\label{table:results_graph}
\begin{tabular}{l l c c c}
\toprule
\textbf{Models} & \textbf{Embeddings} & \textbf{F1} & \textbf{AUROC} & \textbf{Accuracy} \\
\midrule
GraphSAGE    & ProtVec       & 0.7579   & 84.26  & 75.88   \\
         & ESM-2           & 0.7604   & 84.26  & 75.94   \\
         & BioEmbedding & \textbf{0.7896}   & \textbf{87.37}  & \textbf{79.16}   \\
         & AlphaFold2     & 0.7616   & 84.47  & 76.17   \\
\midrule
GAT      & ProtVec       & 0.7348   & 80.64  & 72.89   \\
         & ESM-2          & 0.7313   & 80.35  & 72.52   \\
         & BioEmbedding & \textbf{0.7503}   & \textbf{83.71}  & \textbf{75.43}   \\
         & AlphaFold2     & 0.7401   & 81.35  & 73.44   \\
\midrule
GCN      & ProtVec       & 0.6419   & 69.60  & 63.15   \\
         & ESM-2          & \textbf{0.6620}   & 70.56  & 64.07   \\
         & BioEmbedding & 0.6489   & \textbf{71.00}  & \textbf{64.43}   \\
         & AlphaFold2     & 0.6515   & 69.51  & 63.11   \\
\midrule
GTN      & ProtVec       & 0.6876   & 73.94  & 67.07   \\
         & ESM-2          & 0.6872   & 72.63  & 65.90   \\
         & BioEmbedding & \textbf{0.7154}   & \textbf{79.32}  & \textbf{71.53}   \\
         & AlphaFold2     & 0.6935   & 74.58  & 67.71   \\
\midrule
GIN      & ProtVec       & 0.6495   & 70.31  & 63.90   \\
         & ESM-2           & \textbf{0.6543}   & \textbf{71.09}  & 64.44   \\
         & BioEmbedding & 0.6382   & 68.84  & \textbf{65.21}   \\
         & AlphaFold2     & 0.6392   & 70.37  & 63.81   \\
\bottomrule
\end{tabular}
\end{table}    
\vspace{-1em}  

%
%
\subsubsection{Performance Analysis and Biological Significance}

We evaluated the models using F1-score, AUROC, and accuracy.
Each metric was computed on the held-out test set using threshold-optimized binary predictions derived from predicted logits.
\tableref{results_graph} summarizes the results for each model-embedding combination.
GraphSAGE with BioEmbedding yielded the highest AUROC of 87.37\%, indicating the strongest ability in distinguishing between  interacting versus non-interacting protein pairs.
This configuration also achieved the best F1-score (0.7896) and overall accuracy (79.16\%), outperforming even structure-aware embeddings such as AlphaFold2.

GAT also benefited from BioEmbedding, achieving an AUROC of 83.71\%.
GTN performed competitively with AUROC values of 79.32\% (BioEmbedding) and 74.58\% (AlphaFold2).
GCN consistently showed the weakest performance across all embedding types, likely due to its limited ability to capture higher-order neighborhood dependencies.

GIN showed moderate performance with AUROC around 70\% on most embeddings, slightly underperforming GTN and GraphSAGE.
Despite incorporating edge weight, GIN demonstrated suboptimal performance in the PPI task, likely due to its lack of attention mechanisms or adaptive neighborhood sampling strategies.
Across all models, BioEmbedding consistently provided the most informative features, outperforming both unsupervised sequence models (ProtVec, ESM-2) and structure-aware embeddings (AlphaFold2).
In contrast to AlphaFold2's structural focus, BioEmbedding captures sequence-derived functional properties such as evolutionary context, remote homology, and sequence-order dependencies.
This makes BioEmbedding particularly effective for PPI prediction.

The generated predictions can act as preliminary hypotheses for wet-lab validation, enabling a cost-effective and scalable strategy to investigate protein relationships.
BioEmbedding-based models capture functionally meaningful signals, making them especially valuable for understanding how protein interactions in pathways, complexes, disease contexts, and drug discovery.

%
%
\subsection{Use Case 2: Node Classification for Enzyme Function Prediction  via Semi-Supervised Learning}

We performed binary classification to distinguish enzymes from non-enzymes using different protein embeddings from \dataset.
The workflow used two sequential stages: unsupervised representation learning with a GCN, followed by supervised evaluation with logistic regression.

%
%
\subsubsection{Representation Learning and Classification Methodology}

\textbf{Stage 1: Unsupervised Node Representation Learning.}
We constructed a PPI graph with 21,435 human proteins.
Each node used one of five embedding types: AlphaFold2, ProtVec, ESM-2, BioEmbedding, or a concatenated vector (combining all four).
All embeddings were L2-normalized except ESM-2, which is already normalized.
Proteins were labeled as enzyme or non-enzyme based on UniProt annotations.
The dataset contained 16,675 non-enzyme proteins and 4,660 enzyme proteins, creating a moderate class imbalance that was addressed using class-weighted logistic regression during evaluation.
We trained a three-layer GCN using self-supervised contrastive learning with \texttt{CosineEmbeddingLoss}.
Positive pairs were from PPIs on the graph; negative pairs were randomly sampled uniformly.
The architecture included batch normalization, GELU activations, 30\% dropout, and final projection to a 32-dimensional embedding space.
We trained for 200 epochs utilizing the Adam optimizer with the  learning rate being \(10^{-2}\).

\textbf{Stage 2: Supervised Enzyme Classification.}
After obtaining node embeddings from the pre-trained GCN, we trained a logistic regression classifier to predict enzymatic status.
The labeled protein dataset was segregated into training and testing sets with 80\% for training and 20\% for testing.
We applied class balancing and feature standardization for fair evaluation.
This evaluation quantified how well each embedding type preserved functionally relevant information in the unsupervised setting.

\begin{table}[t!]
\centering
\caption{Performance comparison of different protein embeddings on enzyme vs. non-enzyme classification using unsupervised GCN-based node embeddings followed by logistic regression. 
The baseline uses constant node features without biological embeddings.}
\begin{tabular}{lcccc}
\hline
\textbf{Embedding} & \textbf{Accuracy} & \textbf{Precision} & \textbf{Recall} & \textbf{F1-score} \\
\hline
\textit{Baseline } & 0.6174 & 0.7615 & 0.6174 & 0.6509 \\

AlphaFold2     & 0.7331 & 0.7869 & 0.7331 & 0.7506 \\
ProtVec        & 0.6996 & 0.7781 & 0.6996 & 0.7227 \\
ESM-2          & 0.7362 & 0.7975 & 0.7362 & 0.7547 \\
BioEmbedding  & \textbf{0.7742} & \textbf{0.8221} & \textbf{0.7742} & \textbf{0.7886} \\
Concatenated   & 0.7635 & 0.8078 & 0.7635 & 0.7778 \\
\hline
\end{tabular}
\label{table:embedding_results}
\end{table}

\subsubsection{Results and Functional Implications}
\tableref{embedding_results} presents the classification results across all five embedding configurations.
As a baseline, we trained the GCN with constant node features to simulate the absence of protein-specific information.
This setup yielded an accuracy of 61.7\%, highlighting the contribution of graph structure alone.
All biological embeddings outperformed this baseline, indicating the importance of protein-level information in enzyme classification.

Among individual embeddings, BioEmbedding achieved the highest classification performance with 77.42\% accuracy, 82.21\% precision, and 78.86\% F1-score.
ESM-2 and AlphaFold2 embeddings also performed well, while ProtVec showed slightly lower performance.
The concatenated embeddings (combining AlphaFold2, ProtVec, ESM-2, and BioEmbedding) achieved strong performance across all metrics (F1 = 77.78\%), indicating that multimodal representations capture complementary biological signals.

This task demonstrates the capacity of \dataset~to facilitate semi-supervised protein function prediction, where unsupervised pre-training is followed by supervised evaluation on specific biological tasks. 
The results indicate that embeddings from structural, sequence-based, and contextual models preserve meaningful biochemical information relevant to enzyme function.
The improvement observed with concatenated embeddings suggests that combining orthogonal views of protein data can enhance functional resolution.
These findings support \dataset~as a versatile and biologically informative resource for graph-based machine learning applications in proteomics.

%
%
\subsection{Use Case 3: Embedding-based Protein Family Clustering and Multi-class Classification}

To investigate the biological relevance and discriminative capacity of AlphaFold2-derived embeddings, we performed a dual analysis involving unsupervised clustering and supervised classification of protein families.
We used proteins from the 17 most frequently occurring human families in our dataset, each containing at least 60 members (total 2,399 proteins).
The goal was to understand whether family-specific structure is encoded in the raw AlphaFold2 embeddings and whether this structure can be enhanced via deep learning-based embedding refinement.

\subsubsection{Unsupervised Clustering and Embedding Refinement}

\textbf{Unsupervised Clustering of Raw AlphaFold2 Embeddings.}
We applied unsupervised dimensionality reduction to visualize the structure of the raw AlphaFold2 embedding space.
We employed t-distributed Stochastic Neighbor Embedding (t-SNE) to map the high-dimensional representations into two-dimensional space.
The K-means clustering divided the proteins into clusters, with the number of clusters being the same as the number of known families.

\begin{figure*}[]
    \centering
    \begin{subfigure}{0.45\textwidth}
        \centering
        \includegraphics[width=\linewidth]{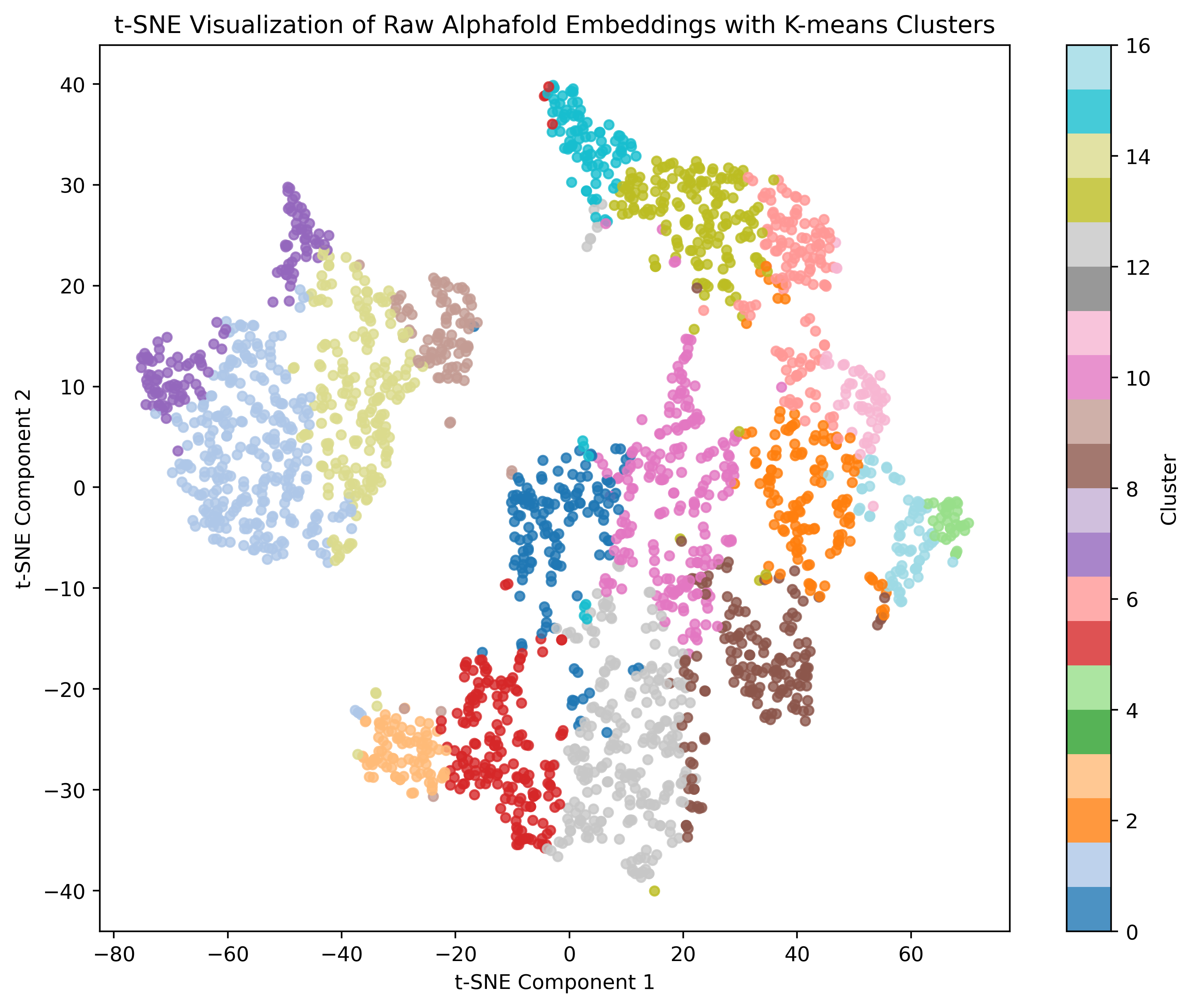}
        \caption{Visualization of raw AlphaFold2 embeddings with K-means clusters. }
        \label{fig:knn_protein_cluster}
    \end{subfigure}
    \hfill
    \begin{subfigure}{0.45\textwidth}
        \centering
        \includegraphics[width=\linewidth]{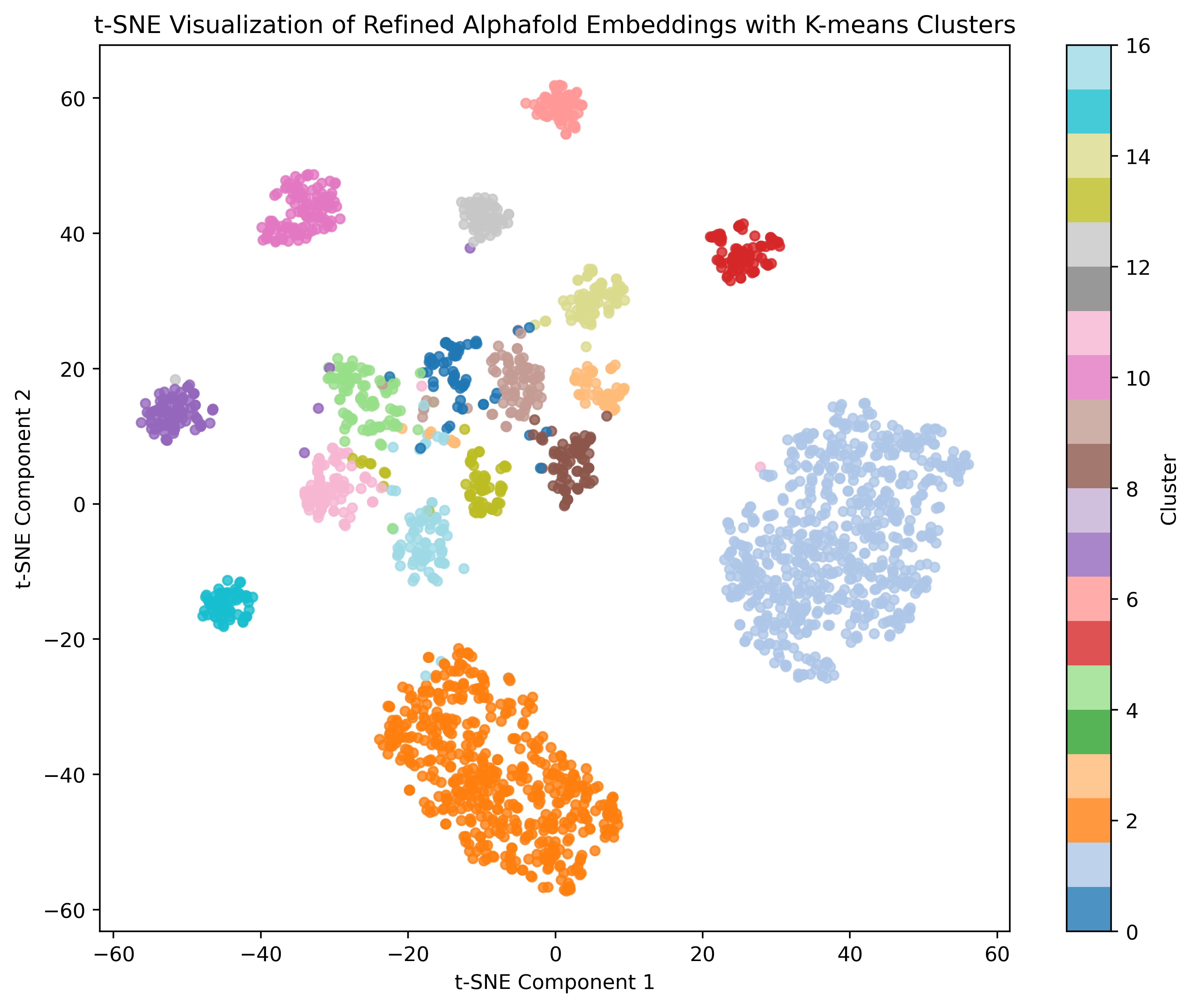}
        \caption{Transformer-refined AlphaFold2 embeddings with K-means clusters.}
        \label{fig:refined_embeddings}
    \end{subfigure}
    \caption{t-SNE comparison of protein embeddings with protein families distinguished by color.
    Each figure shows $2,399$ human proteins from the 17 families listed in \tableref{protein-family}. Additional t-SNE visualizations, including comparisons of raw versus refined embeddings across other modalities are provided in the Supplementary Information.
}
    \label{fig:protein-clusters-comparison}
\end{figure*}

As visualized in \figref{knn_protein_cluster}, some grouping is evident, but many clusters remain diffuse or overlap.
This suggests that while raw AlphaFold2 embeddings carry some family-level information, the boundaries between families are not distinctly separated in the original space.

\textbf{Refinement of AlphaFold2 Embeddings using Transformer Encoder.}
To improve the family separability of the embedding space, we trained a Transformer-based encoder that refines AlphaFold2 embeddings using protein family labels as supervision.
This refinement process reduces noise and redundancy in raw protein representations and emphasizes discriminative features while suppressing less informative dimensions.
This creates a more focused embedding space where evolutionary and functional relationships between protein families become clearer in the t-SNE visualization.

Our methodology employed a LayerNorm and a linear projection from 384 to 128 dimensions, followed by a 2-layer Transformer Encoder (with 4 attention heads and 512-dimensional feedforward networks), and a classification head over 17 family classes.
The network was trained using cross-entropy loss and the Adam optimizer (learning rate $10^{-4}$) for 600 epochs with an 80:20 train-test split and Xavier uniform weight initialization.

After training, we extracted refined 128-dimensional embeddings from the output of the Transformer encoder and re-evaluated their cluster structure using PCA, t-SNE, and K-means clustering.
As shown in Fig.~\ref{fig:protein-clusters-comparison}, the refined embeddings exhibit highly compact and well-separated clusters, showing significantly improved alignment with known family labels.
The model follows the structure described in Table~\ref{tab:transformer_refiner_detailed}.
This marks a substantial improvement over the raw embeddings, demonstrating that the Transformer effectively learns to amplify family-discriminative features embedded in AlphaFold2 representations.

\begin{table*}[htbp]
\centering
\caption{
Transformer encoder architecture used for refining 384-dimensional AlphaFold2 embeddings into 128-dimensional task-specific representations.
The model consists of a normalization layer, a projection into a latent space, two self-attention layers with feed-forward networks, and a classification head.
Final embeddings are extracted before the classifier.}
\label{tab:transformer_refiner_detailed}
\begin{tabular}{p{4cm}p{2cm}p{9.5cm}}

\toprule
\textbf{Layer} & \textbf{Output Dim} & \textbf{Description} \\
\midrule

\texttt{LayerNorm} & $384$ & Layer normalization on input AlphaFold2 embeddings \\
\midrule
\texttt{Linear (Projection)} & $128$ & Fully connected layer; Xavier initialization \\

\midrule
\texttt{Transformer Encoder Layer 1} & $128$ &
\begin{minipage}[t]{\linewidth}
\begin{itemize}
  \item \textbf{Multi-head Self-Attention:} 4 heads, dimension of the key/query vectors, $d_k = 32$ per head
  \item \textbf{Feed-Forward Network:} Linear $\rightarrow$ 512 units $\rightarrow$ GELU $\rightarrow$ Linear $\rightarrow$ 128 units
  \item \textbf{Dropout:} 0.3 (applied after attention and FFN)
  \item \textbf{LayerNorm:} applied after each sub-layer
\end{itemize}
\end{minipage} \\

\midrule
\texttt{Transformer Encoder Layer 2} & $128$ & Same structure as Layer 1 (shared hyperparameters, no weight sharing) \\
\midrule
\texttt{Refined Embedding Output} & $128$ & Final output used for downstream clustering and classifier input \\
\midrule
\texttt{Linear (Classifier Head)} & $17$ & Linear layer mapping to class logits; softmax applied in loss function \\
\bottomrule
\end{tabular}
\end{table*}

\begin{table}[t!]
    \centering
    \caption{Distribution of major human protein families analyzed, showing the $17$ largest families with at least $60$ members each, totaling $2399$ proteins in the dataset.}
    \begin{tabular}{|l|c|}
        \hline
        \multicolumn{1}{|c|}{\textbf{Protein Family}} & \multicolumn{1}{c|}{\textbf{Count}} \\
        \hline
        G-protein coupled receptor 1 family & 681 \\
        krueppel C2H2-type zinc-finger protein family & 537 \\
        peptidase S1 family & 114 \\
        Tyr protein kinase family & 95 \\
        protein-tyrosine phosphatase family & 93 \\
        Rab family & 90 \\
        intermediate filament family & 90 \\
        CMGC Ser/Thr protein kinase family & 81 \\
        CAMK Ser/Thr protein kinase family & 78 \\
        peptidase C19 family & 76 \\
        DEAD box helicase family & 75 \\
        Ser/Thr protein kinase family & 73 \\
        mitochondrial carrier (TC 2.A.29) family & 68 \\
        short-chain dehydrogenases/reductases (SDR)  & 63 \\
        
        STE Ser/Thr protein kinase family & 63 \\
        TRIM/RBCC family & 62 \\
        AGC Ser/Thr protein kinase family & 60 \\
        \hline
         \multicolumn{1}{|r|}{{Total}} & \ 2399 \\
        \hline
    \end{tabular}
    \label{table:protein-family}
\end{table}

We performed unsupervised K-means clustering on both the raw and refined embeddings.
Prior to clustering, the high-dimensional embeddings were reduced using Principal Component Analysis (PCA) to retain 95\% of the variance, a standard practice to denoise and compress representations.
K-means clustering was used with the same number of clusters  as the number of ground-truth families.
Cluster assignments were then mapped to true family labels using majority voting.
Clustering performance was measured using accuracy, precision, recall, F1-score metrics.
Additionally, t-SNE plots were generated to visualize the clustering structure in two dimensions (Fig.~\ref{fig:protein-clusters-comparison}), showing that refined embeddings yield more compact and separable clusters compared to their raw counterparts.

%
%
\subsubsection{Comparative Analysis and Classification Performance}

We considered families containing more than $60$ members of human proteins, resulting in a subset of $2399$ proteins, as shown in \tableref{protein-family}.
As can be observed in \figref{protein-clusters-comparison}, t-SNE 
visualization offered advantages through its preservation of local structure while allowing divergent relationships with greater distances in the reduced-dimensional space.
Unlike Principal Component Analysis (PCA), which uses linear projections that often fail to capture non-linear relationships between protein families, t-SNE effectively revealed the distinct clustering patterns in our refined embeddings.

When comparing \figref{protein-clusters-comparison}(a) versus (b), the refined embeddings showed marked improvement in cluster definition with clearly separated and concentrated data points. 
The Transformer effectively emphasized signals that distinguish protein families while minimizing less relevant variations.
This refinement created embeddings that formed dense clusters with clear boundaries, highlighting stronger intra-family similarities and accentuating inter-family differences.
The density patterns in the refined embedding visualizations provide potential insights into hierarchical relationships between protein families, with proximal clusters often representing evolutionarily or functionally related families.

\begin{table*}[htbp]
\centering
\caption{
Comparison of classification metrics using raw versus refined AlphaFold2 embeddings across multiple models.
Refined embeddings consistently improve F1-score and accuracy across most classifiers.}
\label{tab:raw_vs_refined_embeddings}
\begin{tabular}{l cccc |cccc}
\toprule
\textbf{Model} & \multicolumn{4}{c}{\textbf{Raw Embeddings}} & \multicolumn{4}{c}{\textbf{Refined Embeddings}} \\
\cmidrule(lr){2-5} \cmidrule(lr){6-9}
 & Accuracy & Precision & Recall & F1-Score & Accuracy & Precision & Recall & F1-Score \\
\midrule
K-Means        & 0.6140 & 0.4997 & 0.6140 & 0.5448 & \textbf{0.9621} & \textbf{0.9637} & \textbf{0.9621} & \textbf{0.9622} \\
Naive Bayes    & 0.7562 & 0.7763 & 0.7562 & 0.7591 & 0.8521 & 0.8751 & 0.8521 & 0.8605 \\
Decision Tree  & 0.6979 & 0.7390 & 0.6979 & 0.7109 & 0.7583 & 0.7760 & 0.7583 & 0.7567 \\
FCN            & 0.7688 & 0.7693 & 0.7688 & 0.7517 & 0.8729 & 0.8764 & 0.8729 & 0.8717 \\
KNN            & 0.8000 & 0.8216 & 0.8000 & 0.8048 & 0.8833 & 0.8883 & 0.8833 & 0.8833 \\
Random Forest  & \textbf{0.8438} & \textbf{0.8422} & \textbf{0.8438} & \textbf{0.8393} & 0.8771 & 0.8852 & 0.8771 & 0.8779 \\
\bottomrule
\end{tabular}
\end{table*}

%
%
\textbf{Supervised Protein Family Classification.}
We implemented a supervised multi-class classification pipeline to evaluate the discriminative power of protein embeddings for predicting biological family.
We tested well different types of protein embeddings---both raw and refined---can classify proteins into their correct functional families.

We used a high-quality subset of proteins, filtered to include only families with more than 60 members to ensure balanced representation across classes.
Target labels were derived from curated UniProt annotations, and we applied label encoding to convert string-based family names into numeric class indices for classification.

To benchmark performance, we evaluated classifiers spanning classical machine learning and deep learning approaches.
Each classifier was trained and tested on two embedding types: (i) raw high-dimensional AlphaFold2 embeddings and (ii) refined embeddings from a Transformer encoder trained to enhance class separability.
We standardized all embedding vectors using \texttt{StandardScaler} and used a fixed random seed for 80/20 splits.
Classical classifiers used \texttt{scikit-learn}'s implementations, while neural models used PyTorch with Adam optimizer, dropout, and learning rate scheduling.
The Fully Connected Network (FCN) used five dense layers with GELU activations and batch normalization for stable training.
In contrast, the Transformer used two self-attention layers with a 128-dimensional projection layer.
We evaluated models on the held-out test set using accuracy, precision, recall, and F1-score with weighted averaging to address class imbalance.
Table~\ref{tab:raw_vs_refined_embeddings} shows consistent improvement using refined embeddings.
The highest performance came from Transformer-refined embeddings combined with FCN and KNN.

\textbf{Embedding Refinement Enhances Biological Discrimination.}
Raw AlphaFold2 embeddings, despite being powerful structural representations, are unsupervised and task-agnostic \cite{AlphaFold2}.
They encode broad structural features without optimizing for functional or evolutionary distinctions such as protein family classification.
Although these representations retain biologically relevant information, their discriminative capacity is reduced by task-irrelevant variance. 
Embedding refinement addresses this by introducing supervised signals that guide the latent space toward biologically meaningful organization.
Refined embeddings provide more functionally discriminative representations of proteins, bridging the gap between structure-derived features and functional annotation tasks.

%
%
\section{Discussion and Conclusions}

Understanding and predicting PPIs remains a major challenge in computational biology, largely because molecular relationships are complex and multifaceted.
We introduced \dataset, a human-specific, multimodal PPI resource that integrates diverse embeddings---AlphaFold2, ProtVec, ESM-2, and BioEmbedding---within a unified graph framework.
Table~\ref{tab:use_case_comparison} provides a detailed comparison across three use cases, summarizing their task objectives, methods, embedding types, and key outcomes.

BioEmbeddings consistently performed best across tasks, particularly with GraphSAGE, likely because its transformer-based architecture captures contextual sequence patterns that correlate with protein function.
AlphaFold2 provides structural insights, but its raw embeddings were less effective for link prediction until supervised refinement improved performance in protein family classification.
Concatenated embeddings outperformed individual embeddings in enzyme classification, showing that combining biological features improves results.
GraphSAGE and GTN outperformed GCN, demonstrating that complex models better capture interaction patterns.
These findings establish \dataset~as a powerful and adaptable platform for advancing PPI prediction and protein function analysis.

Across link prediction, enzyme classification, and family clustering, BioEmbeddings consistently provided the most informative features, while structural embeddings required refinement for optimal performance.
This functional context proves essential for protein analysis.
Approaches that integrate both structural and functional information consistently outperform methods relying on structure alone.


The performance differences between model architectures warrant discussion.
The strong performance of GraphSAGE—achieving up to 87.37\% AUROC in PPI prediction—demonstrates the efficacy of neighborhood sampling for capturing informative patterns within biological networks.
GraphSAGE scales efficiently to large graphs while preserving local neighborhood structure, making it well-suited for protein interaction networks.
Graph Transformer Networks also performed strongly, highlighting how attention mechanisms capture the hierarchical and diverse nature of protein relationships.

In contrast, GCN models showed relatively weaker performance (AUROC consistently below 71\%), indicating limitations in modeling complex dependencies in protein interaction networks.
Message-passing limited to immediate neighborhoods fails to capture the multi-scale and hierarchical patterns in biological systems.
Hybrid architectures that combine GraphSAGE’s scalable neighborhood sampling with attention mechanisms may further improve accuracy.
Such approaches may enable more accurate and biologically grounded protein interaction analyses.

Protein family classification shows how refinement improves embeddings.
Supervised fine-tuning increased classification accuracy to 86.04\%, proving that task-specific training enhances structural embeddings for biological tasks.
Iterative, task-specific refinement could improve protein embeddings for drug target prediction and pathway analysis.

\begin{table*}[htbp]
\centering
\caption{Comparison of three benchmark use cases supported by DPEB: task objectives, methods, models, embeddings, and key findings.}
\label{tab:use_case_comparison}
\resizebox{\textwidth}{!}{

\begin{tabular}{p{3.5cm} p{3cm} p{4cm} p{3cm} p{5.5cm}}
\toprule
\textbf{Use Case} & \textbf{Task} & \textbf{Approach} & \textbf{Embeddings Used} & \textbf{Key Findings} \\
\midrule

\textbf{Use Case 1: Supervised PPI Link Prediction} & Predict whether a pair of proteins interact & Supervised edge classification using GNNs with binary cross-entropy loss & AlphaFold2, ProtVec, ESM-2, BioEmbedding & BioEmbedding with GraphSAGE achieved highest AUROC (87.37\%), F1-score (0.7896), and accuracy (79.16\%). Structure-based embeddings like AlphaFold2 were less effective. \\ 

\textbf{Use Case 2: Semi-Supervised Enzyme Classification} & Classify proteins as enzymes or non-enzymes & Self-supervised GCN to learn node embeddings followed by logistic regression & AlphaFold2, ProtVec, ESM-2, BioEmbedding, Concatenated & BioEmbedding achieved highest accuracy (77.42\%). Concatenated embeddings also performed well, proving that different embeddings complement each other. \\ 

\textbf{Use Case 3: Protein Family Clustering and Classification} & Group proteins by functional family & Transformer-based embedding refinement followed by t-SNE+KMeans clustering and multi-class classification & AlphaFold2 (raw and refined) & Refined AlphaFold2 embeddings showed clear cluster separation and improved classification (F1-score: 0.9622 with KMeans, 87\%+ with FCN/KNN).\\

\bottomrule
\end{tabular}
}
\end{table*}

A key contribution of this work is demonstrating how different embedding approaches complement each other.
Multimodal representations outperformed single embeddings, particularly in enzyme classification.
This indicates that individual embedding types cannot fully capture protein complexity.
Future methods should integrate sequence, structural, and functional information for better protein representation.

Several limitations of \dataset~should be acknowledged.
Our focus on human proteins, while providing domain specificity, limits cross-species applicability.
The static nature of pre-computed embeddings limits their ability to reflect context-dependent protein properties, particularly those that vary across different cellular states or environmental conditions.
Dynamic embedding approaches that adjust representations based on biological context are an important direction for future research.

\dataset~makes several design decisions that differentiate it from existing protein databases.
By combining embedding diversity with graph-based learning methods, we provide researchers with flexibility in modeling protein relationships.
A key strength of \dataset~lies in its embedding access framework (see the \textit{Data Availability} section), allowing researchers to retrieve different embeddings by UniProt IDs and enabling integration into downstream analyses.

Future development of \dataset~will focus on expanding coverage to include more human proteins, incorporating new embedding techniques as they emerge, and developing architectures that leverage multimodal protein representations more effectively.
We plan to extend the framework to enable dynamic updating of protein embeddings as new structural and functional data becomes available.

\dataset~serves as a resource for advancing PPI research.
Its scalability, multimodal approach, and design enable applications in systems biology, drug discovery, and personalized medicine.

\section{Data Availability}


\begin{figure*}[t!]
    \centering
    \includegraphics[width=0.8\textwidth]{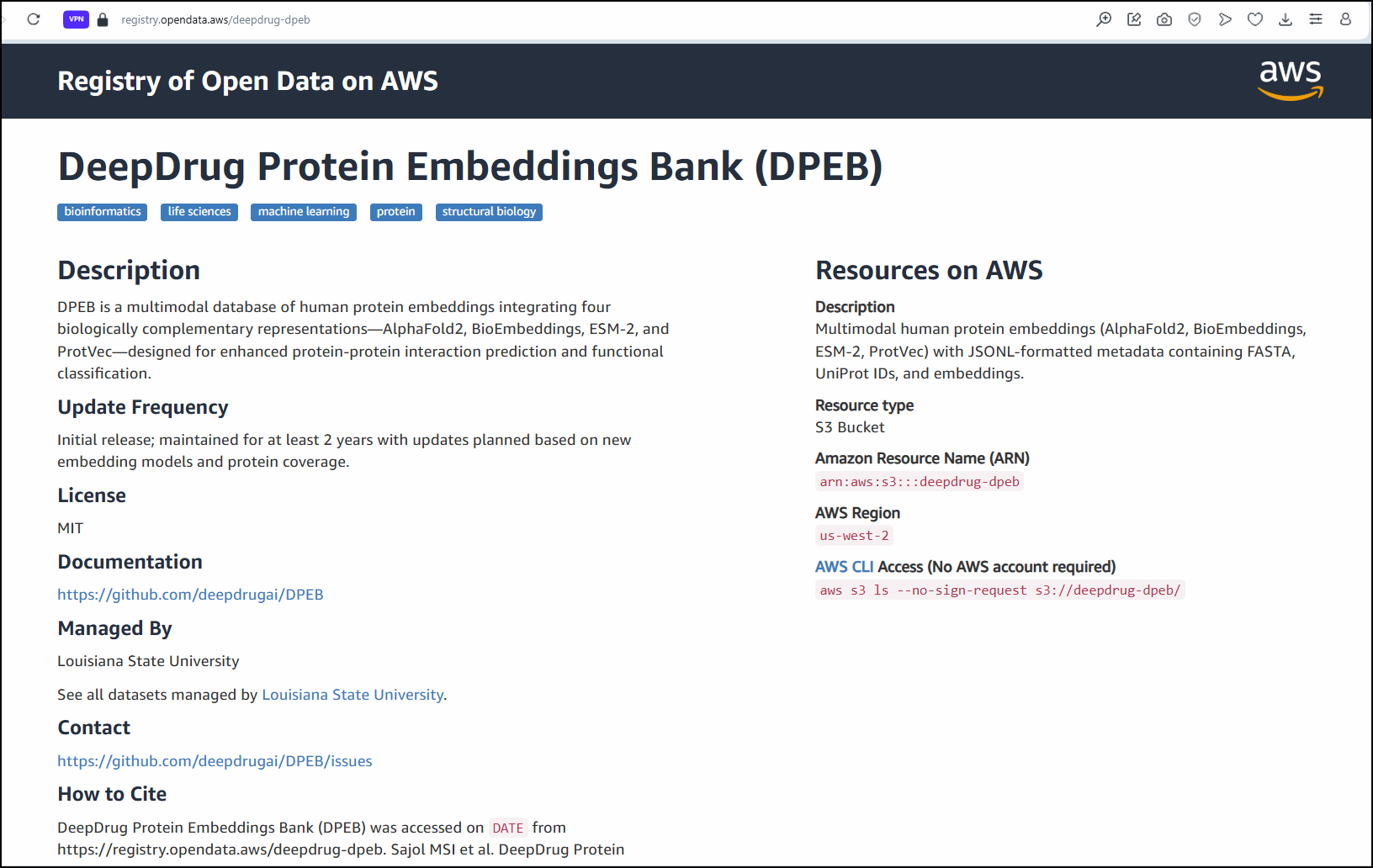}
  \caption{Snapshot of AWS Open Data page for the DeepDrug Protein Embeddings Bank (DPEB) dataset. 
The resource provides multimodal human protein embeddings (AlphaFold2, BioEmbeddings, ESM-2, ProtVec) 
with associated metadata, hosted under the AWS Open Data Sponsorship Program. 
Available at: \url{https://registry.opendata.aws/deepdrug-dpeb/}}

    \label{fig:aws_dpeb}
\end{figure*}

 \dataset~ is publicly available through the AWS Open Data Program (\href{https://registry.opendata.aws/deepdrug-dpeb/}{\textcolor{blue}{https://registry.opendata.aws/deepdrug-dpeb/}}), which is an Amazon Web Services initiative that supports free, large-scale, and long-term hosting of open scientific datasets. 
 The database is designed to support diverse downstream applications such as protein–protein interaction prediction, enzyme classification, and protein family analysis and so on. 
The source code and scripts for reproducing the experiments in this paper are available at: 
\href{https://github.com/deepdrugai/DPEB}{\textcolor{blue}{https://github.com/deepdrugai/DPEB}}.
Detailed documentation of the file structure and data usage example is provided in the linked GitHub repository as well as Supplementary Information section.

The resource will be actively maintained  with updates planned as new embedding models and expanded protein coverage become available. We designate the present release of the dataset as 
\dataset~v1.0.0. 
All files corresponding to this release are permanently archived in the versioned directory 
\texttt{s3://deepdrug-dpeb/v1.0.0/}. 
Future updates will be deposited under incremented version labels (e.g., v1.1.0, v2.0.0), ensuring that past releases remain accessible for reproducibility of published analyses.

The dataset is released under the Creative Commons Attribution 4.0 International License 
(\href{https://creativecommons.org/licenses/by/4.0/}{CC-BY 4.0}), permitting unrestricted use, distribution, 
and reproduction in any medium, provided the original work is properly cited. 
Source codes used for data processing and analysis is released under the MIT License 
(\href{https://opensource.org/licenses/MIT}{MIT}), ensuring free and open use for both academic and industrial research.

\section{Supplementary Data}
The Supplementary Information includes extended results—such as bootstrap robustness analyses, McNemar tests, and clustering/classification visualizations—as well as detailed tables for enzyme and protein family classification. It also provides dataset organization, file descriptions, and programmatic access instructions for the DeepDrug Protein Embeddings Bank (DPEB).

\section*{Acknowledgements}

This dataset is hosted via the \href{https://registry.opendata.aws/open-data/}{\textcolor{black}{Amazon Web Services (AWS) Open Data Sponsorship Program}}.
We gratefully acknowledge AWS for providing cloud storage and bandwidth support, which ensures that DPEB remains freely accessible to the research community.  




\bibliographystyle{plain}
\bibliography{references}

\begin{thebibliography}{10}

\bibitem{alanis2017}
Gregorio Alanis-Lobato, Miguel~A Andrade-Navarro, and Martin~H Schaefer.
\newblock Hippie v2. 0: enhancing meaningfulness and reliability of protein--protein interaction networks.
\newblock {\em Nucleic acids research}, 45(D1):D408--D414, 2017.

\bibitem{alonso2016}
Diego Alonso-L{\'o}pez, Miguel~A Guti{\'e}rrez, Kiran~P Lopes, Carlos Prieto, Ramon Santamar{\'\i}a, and Javier De~Las~Rivas.
\newblock Apid interactomes: providing proteome-based interactomes with controlled quality for multiple species and derived networks.
\newblock {\em Nucleic acids research}, 44(W1):W529--W535, 2016.

\bibitem{protvec}
Ehsaneddin Asgari and Mohammad R.~K. Mofrad.
\newblock Continuous distributed representation of biological sequences for deep proteomics and genomics.
\newblock {\em PLOS ONE}, 10(11):1--15, 11 2015.

\bibitem{bernata2024pcmol}
Andrius Bernatavicius, Martin \v{S}\'{i}cho, Antonius P.~A. Janssen, Alan~Kai Hassen, Mike Preuss, and Gerard J.~P. van Westen.
\newblock Alphafold meets de novo drug design: Leveraging structural protein information in multitarget molecular generative models.
\newblock {\em Journal of Chemical Information and Modeling}, 64(21):8113--8122, 2024.

\bibitem{chen2019pipr}
Muhao Chen, Chelsea J~T Ju, Guangyu Zhou, Xuelu Chen, Tianran Zhang, Kai-Wei Chang, Carlo Zaniolo, and Wei Wang.
\newblock Multifaceted protein–protein interaction prediction based on siamese residual rcnn.
\newblock {\em Bioinformatics}, 35(14):i305--i314, 07 2019.

\bibitem{Chen2019}
Muhao Chen, Chelsea J~T Ju, Guangyu Zhou, Xuelu Chen, Tianran Zhang, Kai-Wei Chang, Carlo Zaniolo, and Wei Wang.
\newblock Multifaceted protein–protein interaction prediction based on siamese residual rcnn.
\newblock {\em Bioinformatics}, 35(14):i305–i314, July 2019.

\bibitem{uniprot}
The~UniProt Consortium.
\newblock Uniprot: the universal protein knowledgebase in 2021.
\newblock {\em Nucleic Acids Research}, 49(D1):D480--D489, 11 2020.

\bibitem{bioembed}
Christian Dallago, Konstantin Schütze, Michael Heinzinger, Tobias Olenyi, Maria Littmann, Amy~X Lu, Kevin~K Yang, Seonwoo Min, Sungroh Yoon, James~T Morton, and Burkhard Rost.
\newblock Learned embeddings from deep learning to visualize and predict protein sets.
\newblock {\em Current Protocols}, 1(5):e113, May 2021.

\bibitem{pcpip}
{Das, S.} and {Chakrabarti, S.}
\newblock Classification and prediction of protein–protein interaction interface using machine learning algorithm.
\newblock {\em Scientific Reports}, 11(1761), 2021.

\bibitem{humap}
Kevin Drew, John~B Wallingford, and Edward~M Marcotte.
\newblock hu.map 2.0: integration of over 15,000 proteomic experiments builds a global compendium of human multiprotein assemblies.
\newblock {\em Molecular Systems Biology}, 17:e10016, 2021.

\bibitem{elnaggar2021prottranscrackinglanguagelifes}
Ahmed Elnaggar, Michael Heinzinger, Christian Dallago, Ghalia Rihawi, Yu~Wang, Llion Jones, Tom Gibbs, Tamas Feher, Christoph Angerer, Martin Steinegger, Debsindhu Bhowmik, and Burkhard Rost.
\newblock Prottrans: Towards cracking the language of life's code through self-supervised deep learning and high performance computing, 2021.

\bibitem{guan2025graphrpi}
Jiahui Guan, Lantian Yao, Peilin Xie, Zhihao Zhao, Dian Meng, Tzong-Yi Lee, Junwen Wang, and Ying-Chih Chiang.
\newblock Graph-rpi: predicting rna--protein interactions via graph autoencoder and self-supervised learning strategies.
\newblock {\em Brief Bioinform}, 26(3):bbaf292, 2025.

\bibitem{acc_svm}
Yanzhi Guo, Lezheng Yu, Zhining Wen, and Menglong Li.
\newblock {Using support vector machine combined with auto covariance to predict protein–protein interactions from protein sequences}.
\newblock {\em Nucleic Acids Research}, 36(9):3025--3030, 04 2008.

\bibitem{massa}
Fan Hu, Yishen Hu, Weihong Zhang, Huazhen Huang, Yi~Pan, and Peng Yin.
\newblock A multimodal protein representation framework for quantifying transferability across biochemical downstream tasks.
\newblock {\em Advanced Science}, page 2301223, 2023.

\bibitem{huttlin2017}
Edward~L Huttlin, Raphael~J Bruckner, Joao~A Paulo, Joe~R Cannon, Lily Ting, Kristin Baltier, Greg Colby, Fana Gebreab, Melanie~P Gygi, Hannah Parzen, et~al.
\newblock Architecture of the human interactome defines protein communities and disease networks.
\newblock {\em Nature}, 545(7655):505--509, 2017.

\bibitem{huttlin2015}
Edward~L Huttlin, Lily Ting, Raphael~J Bruckner, Fana Gebreab, Melanie~P Gygi, John Szpyt, Stanley Tam, Gabriela Zarraga, Greg Colby, Kristin Baltier, et~al.
\newblock The bioplex network: a systematic exploration of the human interactome.
\newblock {\em Cell}, 162(2):425--440, 2015.

\bibitem{jha2022ppi}
K.~Jha, S.~Saha, and H.~Singh.
\newblock Prediction of protein--protein interaction using graph neural networks.
\newblock {\em Sci Rep}, 12:8360, 2022.

\bibitem{AlphaFold2}
John Jumper, Richard Evans, Alexander Pritzel, Tim Green, Michael Figurnov, Olaf Ronneberger, Kathryn Tunyasuvunakool, Russ Bates, Augustin Zidek, Anna Potapenko, et~al.
\newblock Highly accurate protein structure prediction with alphafold.
\newblock {\em Nature}, 596(7873):583--589, 2021.

\bibitem{Kamp2025GNNsuite}
Sebesty{\'e}n Kamp, Giovanni Stracquadanio, and T.~Ian Simpson.
\newblock {GNN-Suite}: A graph neural network bench-marking framework for biomedical informatics.
\newblock {\em arXiv preprint arXiv:2505.10711}, 2025.
\newblock Comparative framework includes GCN, GAT, GraphSAGE, GIN, GTN, etc. for PPI networks:contentReference[oaicite:12]{index=12}.

\bibitem{keshava2009}
T.~S. Keshava~Prasad, Renu Goel, Kumaran Kandasamy, Shivakumar Keerthikumar, Sameer Kumar, Suresh Mathivanan, Deepthi Telikicherla, Rajesh Raju, Beema Shafreen, Abhilash Venugopal, et~al.
\newblock Human protein reference database—2009 update.
\newblock {\em Nucleic acids research}, 37(suppl\_1):D767--D772, 2009.

\bibitem{kotlyar2019}
Max Kotlyar, Chiara Pastrello, Zara Malik, and Igor Jurisica.
\newblock Iid 2018 update: context-specific physical protein--protein interactions in human, model organisms and domesticated species.
\newblock {\em Nucleic acids research}, 47(D1):D581--D589, 2019.

\bibitem{hang2018dnn}
Hang Li, Xiu-Jun Gong, Hua Yu, and Chang Zhou.
\newblock Deep neural network based predictions of protein interactions using primary sequences.
\newblock {\em Molecules}, 23(8):1923, 2018.

\bibitem{esm}
Zeming Lin, Halil Akin, Roshan Rao, Brian Hie, Zhongkai Zhu, Wenting Lu, Nikita Smetanin, Robert Verkuil, Ori Kabeli, Yaniv Shmueli, Allan dos Santos~Costa, Maryam Fazel-Zarandi, Tom Sercu, Salvatore Candido, and Alexander Rives.
\newblock Evolutionary-scale prediction of atomic-level protein structure with a language model.
\newblock {\em Science}, 379(6637):1123--1130, 2023.

\bibitem{lv2021gnnppi}
Guofeng Lv, Zhiqiang Hu, Yanguang Bi, and Shaoting Zhang.
\newblock Learning unknown from correlations: Graph neural network for inter-novel-protein interaction prediction.
\newblock {\em CoRR}, abs/2105.06709, 2021.

\bibitem{Ma2024Bingo}
Jiani Ma, Jiangning Song, Neil~D. Young, Bill C.~H. Chang, Pasi~K. Korhonen, Tulio~L. Campos, Hui Liu, and Robin~B. Gasser.
\newblock {'Bingo'}: a large language model- and graph neural network-based workflow for predicting essential genes from protein data.
\newblock {\em Briefings in Bioinformatics}, 25(1):bbad472, 2024.
\newblock Compares GCN, GAT, GraphSAGE, GIN on PPI data:contentReference[oaicite:13]{index=13}.

\bibitem{word2vec}
Tom{\'{a}}s Mikolov, Ilya Sutskever, Kai Chen, Gregory~S. Corrado, and Jeffrey Dean.
\newblock Distributed representations of words and phrases and their compositionality.
\newblock In Christopher J.~C. Burges, L{\'{e}}on Bottou, Zoubin Ghahramani, and Kilian~Q. Weinberger, editors, {\em Advances in Neural Information Processing Systems 26: 27th Annual Conference on Neural Information Processing Systems 2013. Proceedings of a meeting held December 5-8, 2013, Lake Tahoe, Nevada, United States}, pages 3111--3119, 2013.

\bibitem{Neumann2022}
Don Neumann, Soumyadip Roy, Fayyaz Ul Amir~Afsar Minhas, and Asa Ben-Hur.
\newblock On the choice of negative examples for prediction of host-pathogen protein interactions.
\newblock {\em Frontiers in Bioinformatics}, 2, December 2022.

\bibitem{intact}
Sandra Orchard, Mais Ammari, Bruno Aranda, Lionel Breuza, Leonardo Briganti, Fiona Broackes-Carter, Nancy~H Campbell, Gayatri Chavali, Carol Chen, Noemi del Toro, Margaret Duesbury, Marine Dumousseau, Eugenia Galeota, Ursula Hinz, Marta Iannuccelli, Sruthi Jagannathan, Rafael Jimenez, Jyoti Khadake, Astrid Lagreid, Luana Licata, Ruth~C Lovering, Birgit Meldal, Anna~N Melidoni, Mila Milagros, Daniele Peluso, Livia Perfetto, Pablo Porras, Arathi Raghunath, Sylvie Ricard-Blum, Bernd Roechert, Andre Stutz, Michael Tognolli, Kim van Roey, Gianni Cesareni, and Henning Hermjakob.
\newblock The mintact project--intact as a common curation platform for 11 molecular interaction databases.
\newblock {\em Nucleic acids research}, 42(Database issue):D358—63, January 2014.

\bibitem{salwinski2004}
Lukasz Salwinski, Christopher~S Miller, Adam~J Smith, Frank~K Pettit, James~U Bowie, and David Eisenberg.
\newblock The database of interacting proteins: 2004 update.
\newblock {\em Nucleic acids research}, 32(suppl\_1):D449--D451, 2004.

\bibitem{song2022tagppi}
Bosheng Song, Xiaoyan Luo, Xiaoli Luo, Yuansheng Liu, Zhangming Niu, and Xiangxiang Zeng.
\newblock Learning spatial structures of proteins improves protein–protein interaction prediction.
\newblock {\em Briefings in Bioinformatics}, 23(2):bbab558, 01 2022.

\bibitem{stark2006}
Chris Stark, Bobby-Joe Breitkreutz, Teresa Reguly, Lorrie Boucher, Ashton Breitkreutz, and Mike Tyers.
\newblock Biogrid: a general repository for interaction datasets.
\newblock {\em Nucleic acids research}, 34(suppl\_1):D535--D539, 2006.

\bibitem{steinegger2018clustering}
Martin Steinegger and Johannes S{\"o}ding.
\newblock Clustering huge protein sequence sets in linear time.
\newblock {\em Nature Communications}, 9(1):2542, 2018.

\bibitem{string}
Damian Szklarczyk, Rebecca Kirsch, Mikaela Koutrouli, Katerina Nastou, Farrokh Mehryary, Radja Hachilif, Annika~L Gable, Tao Fang, Nadezhda~T Doncheva, Sampo Pyysalo, Peer Bork, Lars~J Jensen, and Christian von Mering.
\newblock The string database in 2023: protein-protein association networks and functional enrichment analyses for any sequenced genome of interest.
\newblock {\em Nucleic Acids Research}, 51(D1):D638--D646, Jan 2023.

\bibitem{wu2023dlppi}
Jiahui Wu, Bo~Liu, Jidong Zhang, Zhihan Wang, and Jianqiang Li.
\newblock Dl-ppi: a method on prediction of sequenced protein–protein interaction based on deep learning.
\newblock {\em BMC Bioinformatics}, 24(1):473, 2023.

\bibitem{Wu2023}
Jiahui Wu, Bo~Liu, Jidong Zhang, Zhihan Wang, and Jianqiang Li.
\newblock Dl-ppi: a method on prediction of sequenced protein–protein interaction based on deep learning.
\newblock {\em BMC Bioinformatics}, 24(1), December 2023.

\bibitem{Xu2024SurveyPPIs}
Mingda Xu, Peisheng Qian, Ziyuan Zhao, Zeng Zeng, Jianguo Chen, Weide Liu, and Xulei Yang.
\newblock Graph neural networks for protein-protein interactions -- a short survey.
\newblock {\em arXiv preprint arXiv:2404.10450}, 2024.
\newblock Surveys GNN-based PPI methods; highlights GCN-based (e.g. GraphSAGE, GIN) and GAT-based model families:contentReference[oaicite:14]{index=14}.

\bibitem{code4_knn}
{Yang Lei}, {Xia Jun-Feng}, and {Gui Jie}.
\newblock Prediction of protein-protein interactions from protein sequence using local descriptors.
\newblock {\em Protein \& Peptide Letters}, 17(9), 2010.

\bibitem{mcd_svm}
{You, ZH.}, {Zhu, L.}, {Zheng, CH.}, and {et al.}
\newblock Prediction of protein-protein interactions from amino acid sequences using a novel multi-scale continuous and discontinuous feature set.
\newblock {\em BMC Bioinformatics}, 15 (Suppl 15)(S9), 2014.

\bibitem{gcforest}
Bin Yu, Cheng Chen, Xiaolin Wang, Zhaomin Yu, Anjun Ma, and Bingqiang Liu.
\newblock Prediction of protein–protein interactions based on elastic net and deep forest.
\newblock {\em Expert Systems with Applications}, 176:114876, 2021.

\bibitem{gnngl_ppi}
Xin Zeng, Fan-Fang Meng, Meng-Liang Wen, Shu-Juan Li, and Yi~Li.
\newblock Gnngl-ppi: multi-category prediction of protein-protein interactions using graph neural networks based on global graphs and local subgraphs.
\newblock {\em BMC Genomics}, 25(1):406, 2024.

\bibitem{Zeng2024}
Xin Zeng, Fan-Fang Meng, Meng-Liang Wen, Shu-Juan Li, and Yi~Li.
\newblock Gnngl-ppi: multi-category prediction of protein-protein interactions using graph neural networks based on global graphs and local subgraphs.
\newblock {\em BMC Genomics}, 25(1), May 2024.

\bibitem{Zhou2022ComparativePPI}
Hang Zhou, Weikun Wang, Jiayun Jin, Zengwei Zheng, and Binbin Zhou.
\newblock Graph neural network for protein--protein interaction prediction: A comparative study.
\newblock {\em Molecules}, 27(18):6135, 2022.
\newblock Benchmarks GCN and GAT (among others) as widely-used GNN models for PPI prediction:contentReference[oaicite:15]{index=15}.

\bibitem{ld_svm}
Yu~Zhen Zhou, Yun Gao, and Ying~Ying Zheng.
\newblock Prediction of protein-protein interactions using local description of amino acid sequence.
\newblock In Mark Zhou and Honghua Tan, editors, {\em Advances in Computer Science and Education Applications}, pages 254--262, Berlin, Heidelberg, 2011. Springer Berlin Heidelberg.

\bibitem{lra_rf}
{Zhu-Hong You}, {Xiao Li}, and {Keith CC Chan}.
\newblock An improved sequence-based prediction protocol for protein-protein interactions using amino acids substitution matrix and rotation forest ensemble classifiers.
\newblock {\em Neurocomputing}, 228:277--282, 2017.
\newblock Advanced Intelligent Computing: Theory and Applications.

\end{thebibliography}

\end{document}